\documentclass[sn-mathphys,Numbered,iicol]{sn-jnl}

\usepackage{graphicx}%
\usepackage{multirow}%
\usepackage{amsmath,amssymb,amsfonts}%
\usepackage{amsthm}%
\usepackage{mathrsfs}%
\usepackage[title]{appendix}%
\usepackage{xcolor}%
\usepackage{textcomp}%
\usepackage{manyfoot}%
\usepackage{booktabs}%
\usepackage{algorithm}%
\usepackage{algorithmicx}%
\usepackage{algpseudocode}%
\usepackage{listings}%
\usepackage{cleveref}%

\begin{document}

\title[Article Title]{
RefreshNet: Learning Multiscale Dynamics through Hierarchical Refreshing}


\author*[1]{\fnm{Junaid} \sur{Farooq}}\email{junaid\_phd017@nitsri.ac.in}

\author[]{\fnm{Danish} \sur{Rafiq$^2$}}\email{danish.rafiq@iust.ac.in}

\author[]{\fnm{Pantelis R.} \sur{Vlachas$^3$}}\email{pvlachas@ethz.ch}

\author[1]{\fnm{Mohammad Abid} \sur{Bazaz}}\email{abid@nitsri.ac.in}

\affil[1]{\orgdiv{Department of Electrical Engineering}, \orgname{National Institute of Technology Srinagar}, \orgaddress{\street{Hazratbal}, \city{Srinagar}, \postcode{190006}, \state{Jammu and Kashmir}, \country{India}}}

\affil[2]{\orgdiv{Department of Electrical Engineering}, \orgname{Islamic University of Science and Technology}, \orgaddress{\street{Awantipora}, \city{Pulwama}, \postcode{192122}, \state{Jammu and Kashmir}, \country{India}}}

\affil[3]{\orgdiv{Institute of Structural Engineering}, \orgname{ETH Zurich}, \orgaddress{\city{Zürich}, \postcode{8093}, \state{Zürich}, \country{Switzerland}}}





\maketitle

\abstract{l}
\begin{abstract}

\textbf{Abstract:}
Forecasting complex system dynamics, particularly for long-term predictions, is persistently hindered by error accumulation and computational burdens.
This study presents RefreshNet, a multiscale framework developed to overcome these challenges, delivering an unprecedented balance between computational efficiency and predictive accuracy.
RefreshNet incorporates convolutional autoencoders to identify a reduced order latent space capturing essential features of the dynamics, and strategically employs multiple recurrent neural network (RNN) blocks operating at varying temporal resolutions within the latent space, thus allowing the capture of latent dynamics at multiple temporal scales.
The unique ``refreshing'' mechanism in RefreshNet allows coarser blocks to reset inputs of finer blocks, effectively controlling and alleviating error accumulation.
This design demonstrates superiority over existing techniques regarding computational efficiency and predictive accuracy, especially in long-term forecasting.
The framework is validated using three benchmark applications: the FitzHugh-Nagumo system, the Reaction-Diffusion equation, and Kuramoto-Sivashinsky dynamics.
RefreshNet significantly outperforms state-of-the-art methods in long-term forecasting accuracy and speed, marking a significant advancement in modeling complex systems and opening new avenues in understanding and predicting their behavior.
\end{abstract}




\section{Introduction}\label{sec1}

Advanced modeling and simulation methods~\cite{zeigler2000theory}, encompassing data-driven approaches and those grounded on first principles, have significantly influenced a wide array of quantitative scientific fields.
This impact is evident in diverse domains including language processing~\cite{jozefowicz2016exploring}, speech modeling~\cite{deng2006structured}, financial analytics~\cite{benninga2014financial}, health informatics~\cite{ravi2016deep}, biology~\cite{alber2019integrating}, and epidemiology~\cite{frauenthal2012mathematical}, all of which have reaped substantial benefits from these technologies.
These methods are transformative in their capacity to forecast outcomes, particularly in areas where extensive experimentation is impractical or too expensive, often due to high infrastructure, equipment, and skilled personnel requirements.
However, a critical challenge remains: the precision of these simulations heavily relies on their ability to comprehensively represent the spatial and temporal scales of the phenomena being studied~\cite{kevrekidis2004equation}.

Multiscale systems involve many interacting scales, exhibiting nonlinear, interconnected dynamics across a broad scale range.
From microscopic to macroscopic levels, the varying granularities at which these systems operate complicate the task of accurate prediction.
A striking example of this challenge can be seen in climate modeling \cite{dunlea2012national}.
Scientists must account for various dynamic processes, from microscale atmospheric interactions to macroscale global climate phenomena.
The system's components are not isolated but interconnected, often resulting in nonlinear and feedback-driven interactions.
These features give rise to emergent properties that are not readily predictable from studying individual components in isolation.
Therefore, forecasting future spatiotemporal data for long-term horizons remains a formidable challenge in the predictive modeling of complex systems.

Research efforts aimed at addressing the challenges of multiscale modeling and simulation have culminated in the development of the equation-free framework (EFF)~\cite{kevrekidis2003equation,kevrekidis2004equation}.
EFF facilitates efficient modeling of multiscale high-dimensional systems by integrating coarse-grained and fine-scale simulations~\cite{laing2010reduced,maulik2021reduced}.
However, this approach is not without its own challenges, such as selecting appropriate degrees of freedom for these representations.

Recently, the Learning Effective Dynamics (LED) approach, an advanced extension of the EFF framework that incorporates cutting-edge machine learning architectures into EFF, has shown enhanced predictability over traditional reduced-order models~\cite{vlachas2022multiscale}.
LED refines the simulation process by operating within a reduced order latent space and then expanding to high-dimensional states through the use of convolutional autoencoders~\cite{vlachas2022multiscale}.
Originally introduced as a nonlinear alternative to principal component analysis \cite{moore1981principal}, autoencoders (AEs) offer powerful capabilities for dimensionality reduction and feature learning.
AEs have been used in the identification of non-linear modes across multiple fields from structural mechanics~\cite{simpson2020use,simpson2021machine}, materials modeling~\cite{he2021deep}, fluid flows~\cite{milano2002neural}, to general dynamical systems~\cite{otto2019linearly}.
By capturing nonlinear relationships within the data, AEs can model complex patterns and structures that may not be captured by linear techniques.
Notably, LED autoencoders have proven to be effective in establishing mappings between fine and coarse-grained representations, thereby enhancing simulation efficiency.
LED has been successfully applied to various systems, from molecular simulations~\cite{vlachas2021accelerated} to high-dimensional fluid flows~\cite{vlachas2022multiscale}.

LED employs Recurrent Neural Networks (RNNs)~\cite{elman1990finding,greff2016lstm} to propagate the latent reduced-order dynamics.
Such machine learning architectures specialized in sequential data have revolutionized the field of multi-variate time series forecasting, from dynamical systems~\cite{pathak2018model,vlachas2018data,chang2019antisymmetricrnn,trischler2016synthesis,wan2018data,kundu2022long,zhang2019deep,vlachas2020backpropagation} to audio and speech processing~\cite{salinas2020deepar,oord2016wavenet, hizlisoy2021music,sak2014long,graves2013speech}, achieving orders of magnitude better performance compared to traditional methods.
Multiple other works, similar to LED, have employed RNNs coupled with autoencoders for spatiotemporal dynamics modeling~\cite{zang2020short,o2022spatio,ding2020interpretable,hasegawa2020cnn,simpson2021machine}.

\begin{figure}
    \centering
    \includegraphics[width=0.4\textwidth]{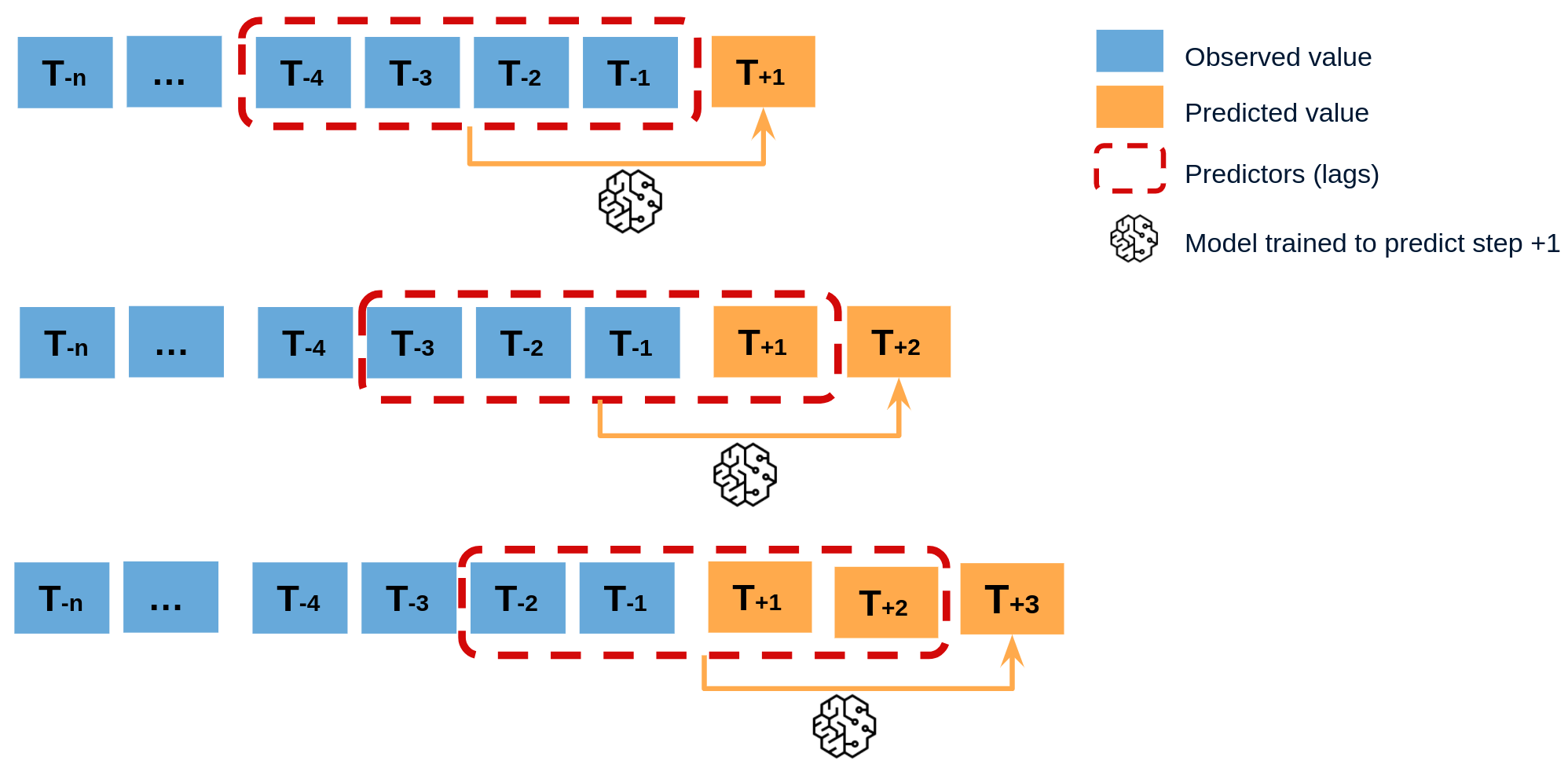}
    \caption{Demonstration of iterative, autoregressive forecasting}
    \label{fig:error_accumulation}
\end{figure}

Autoregressive time series algorithms, like RNNs employed in LED, face a critical shortcoming known as the compounding error effect or error accumulation~\cite{sangiorgio2020robustness,teutsch2022flipped,vlachas2023learning,bengio2009curriculum}.
This phenomenon occurs as these models generate future forecasts based on previous predictions, as depicted in~\Cref{fig:error_accumulation}.
Consequently, errors accumulate at each forecasting step, resulting in diminished forecast quality as we extend further into the future \cite{selim2020reducing}.
This so-called error accumulation problem is widely recognized within the research community, and proposed solutions range from novel architectures that alleviate the autoregressive propagation (e.g., Transformers~\cite{lin2022survey}), to regularization techniques~\cite{alzubaidi2021review}, and novel training methods~\cite{sangiorgio2020robustness,vlachas2023learning,teutsch2022flipped,bengio2009curriculum,brenner2022tractable,hess2023generalized,brenner2022multimodal,durstewitz2023reconstructing}.

Another factor contributing to error accumulation in autoregressive methods is their training objective, which focuses on minimizing the loss in one-step-ahead predictions.
However, during autoregressive testing, these networks base their inferences on self-generated predictions, scenarios not encountered in training.
This phenomenon, known as the exposure bias effect, can negatively impact the RNN's ability to generalize effectively~\cite{peussa2021exposure}.

While LED has demonstrated promising results in the modeling and simulating of multiscale dynamical systems, it continues to utilize autoregressive RNNs for latent space propagation, inheriting associated limitations such as iterative error propagation and exposure bias effect.
The present research is motivated by these persistent challenges in the predictive modeling of complex multiscale systems.
In particular, we address these two critical problems that have yet to be solved: the mismatch between the training objective and autoregressive inference and the pervasive issue of error accumulation in long-term forecasting.

In this paper, we propose a novel framework called RefreshNet, specifically designed to address these challenges by employing a hierarchy of RNNs operating at different timescales within the LED framework.
RefreshNet is designed to address the issue of accumulating errors, providing greater accuracy than LSTMs and LED while also delivering enhanced computational efficiency over the LED framework.
The results of this study promise a leap forward in our ability to forecast the behavior of complex systems over longer horizons, enabling the prediction of the behavior of complex systems across numerous applications.
The unique contributions of our framework include the following:
\begin{enumerate}
\item Error accumulation reduction: RefreshNet incorporates multiple RNN blocks at various temporal resolutions. The coarser blocks refresh the finer blocks' inputs, resetting error accumulation, improving prediction accuracy, and enabling longer prediction times with lower errors.
\item Computational benefits: Our framework has the potential to significantly reduce the computation time by delaying the need to solve equations based on first principles numerically, outperforming state-of-the-art techniques.
\item Multiscale dynamics learning: Integrating multiple temporal resolutions enables effective learning of multiscale dynamics.
\item Scalability: RefreshNet can adapt to complex systems with evolving dynamics by integrating additional RNN blocks at even coarser temporal resolutions.
\end{enumerate}

The remainder of this paper is organized as follows.
\Cref{sec:methods} presents the general methodology of the RefreshNet framework. In \cref{sec:results}, we demonstrate the results of RefreshNet applied on benchmark problems of complex dynamical systems forecasting.
In \cref{sec:discussion}, we discuss the implications, limitations, and future scope of the proposed framework.
Finally, \cref{sec:conclusion} concludes the paper.

\section{Methods}
\label{sec:methods}

The high-dimensional state of a dynamical system is given by $\mathbf{x}_t \in \mathbb{R}^{d_x}$, and the discrete time dynamics are given by
\begin{equation}
\mathbf{x}_{t+\Delta t} = \mathbf{\mathcal{F}}(\mathbf{x}_t)
\end{equation}
where $\Delta t$ is the sampling period and $\mathcal{F}$ may be nonlinear, deterministic or stochastic. 
The state of the system at time $t$ can also be described by a vector $\mathbf{z}_t \in \mathcal{Z}^{d_z}$, where $\mathcal{Z} \subset \mathbb{R}^{d_z}$ is the low dimension manifold with $d_z << d_x$. 
To identify this manifold, we define an encoder $\mathcal{E}^{w_\mathcal{E}} : \mathbb{R}^{d_x} \rightarrow \mathbb{R}^{d_x}$, 
where $w_\mathcal{E}$ represent the trainable parameters. Thus, the high-dimensional state $\mathbf{x}_t$ is transformed to low-dimensional state $\mathbf{z}_t$ such that $\mathbf{z}_t = \mathcal{E}^{w_\mathcal{E}} (\mathbf{x}_t)$. This latent state is mapped back to the original state using a decoder, 
i.e., $\mathbf{\tilde{x}}_t = \mathcal{D}^{w_\mathcal{D}}(\mathbf{z}_t)$.
The combination of the encoder and the decoder are termed together as autoencoder.

\begin{figure}
    \centering
    \includegraphics[width=0.4\textwidth]{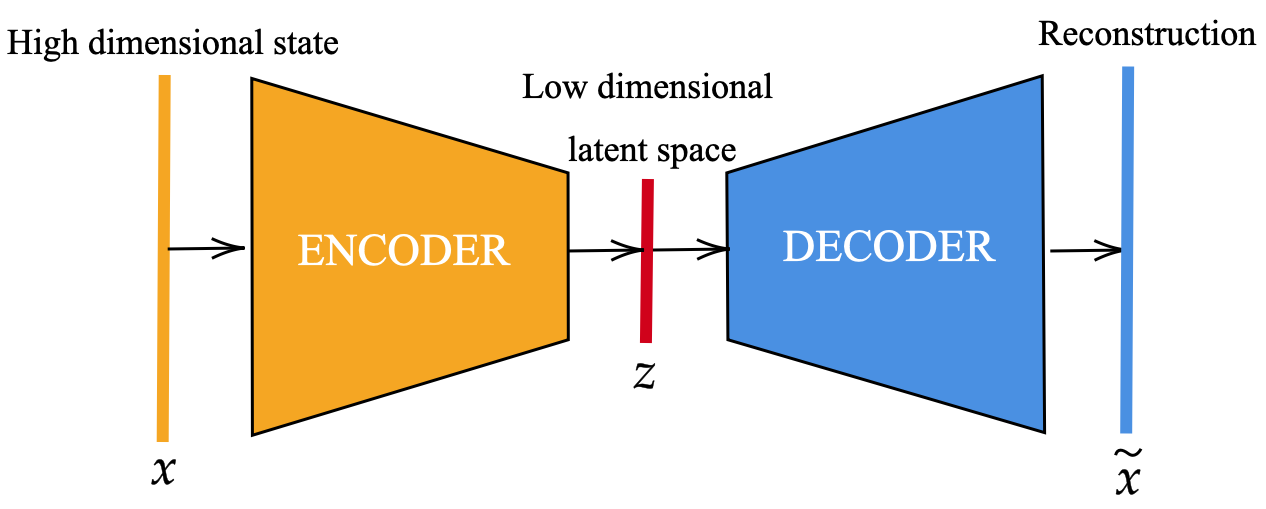}
    \caption{Classical Autoencoder}
    \label{fig:auto_encoder}
\end{figure}

\begin{figure}
    \centering
    \includegraphics[width=0.4\textwidth]{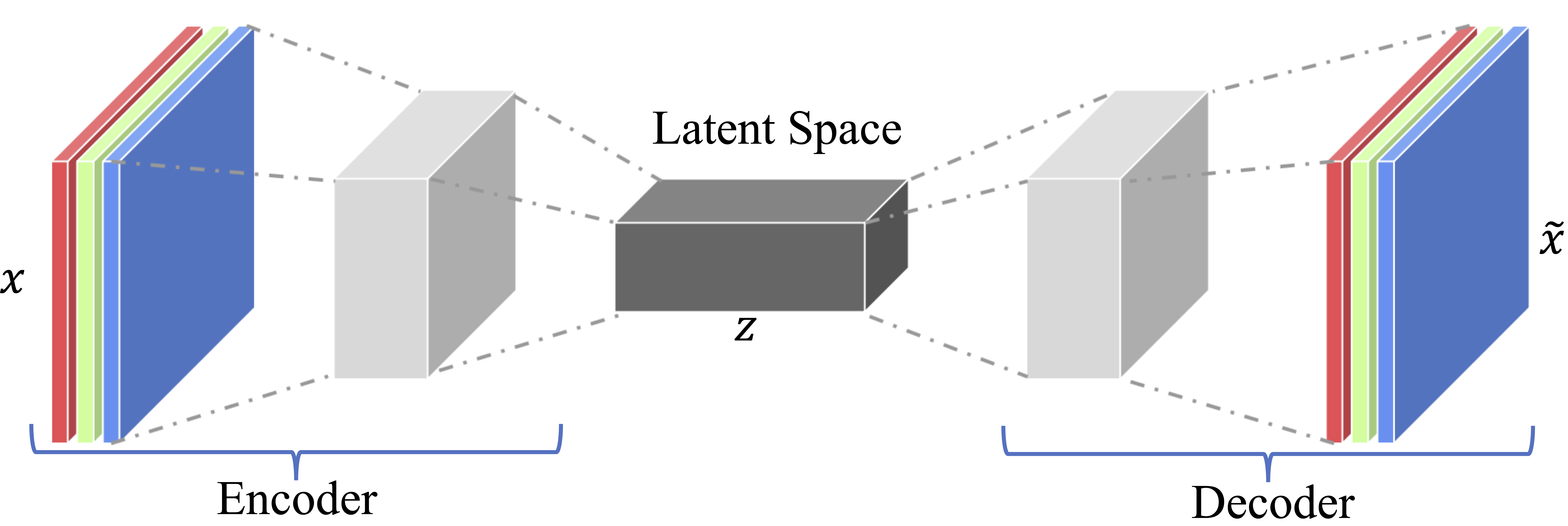}
    \caption{Convolutional autoencoder}
    \label{fig:CAE}
\end{figure}

AEs~\cite{baldi2012autoencoders} are neural networks that utilize nonlinear transformations to map an input to a lower-dimensional latent space and then reconstruct it back to its original dimension at the output.
During training, the objective is to minimize the reconstruction loss, typically measured as the squared difference between the input and the reconstructed output, i.e. $\mathcal{L} = |\mathbf{x} - \mathbf{\tilde{x}}|^2$. 
This loss function guides the AE to learn meaningful representations of the input data.
\Cref{fig:auto_encoder,fig:CAE} provide a visual depiction of an AE, illustrating the flow of information from the input layer to the latent space and back to the output layer.

In this study, we use convolutional neural network (CNN) based AEs.
CNNs have been specifically designed to effectively process image data, leveraging the inherent spatial correlations present in such data.
The architecture of a CNN comprises multiple layers, each  processing a multidimensional input that includes a channel axis and spatial axes.
By employing convolutional kernels or filters that slide along the spatial axes of the input, CNNs exploit the spatial structure of the data.
This characteristic of CNNs can be seen as a geometric prior, as they inherently consider the structural relationships within the data.

In this study, we incorporate CNN layers within an AE framework, where a bottleneck layer is introduced to reduce dimensionality.
The layers commonly employed in a convolutional autoencoder (CAE) include convolutional, pooling, and upsampling layers.
Convolutional layers apply filters to capture spatial features in the input data, while pooling layers downsample the spatial dimensions to reduce computational complexity and extract essential information.
Upsampling layers, also known as transposed convolutional layers or deconvolutional layers, perform the opposite operation of pooling layers by increasing the spatial dimensions of the data.
These layers collectively enable the autoencoder to effectively encode high-dimensional input data into a lower-dimensional latent space and subsequently decode it to reconstruct the original input.
The combination of these layers in a convolutional autoencoder allows for the extraction of meaningful features from complex data and facilitates accurate reconstruction of the input data.
A typical CAE is represented in~\cref{fig:CAE}.

The optimal parameters of the CAE are determined by minimizing the mean squared reconstruction error (MSE):
\begin{equation}
\begin{aligned}
w_\mathcal{E}^*, w_\mathcal{D}^* &= \text{argmin}_{w_\mathcal{E}, w_\mathcal{D}} (\mathbf{x}_t - \mathbf{\tilde{x}}_t)^2 \\
&= \text{argmin}_{w_\mathcal{E}, w_\mathcal{D}} (\mathbf{x}_t - \mathcal{D}^{w_\mathcal{D}} ( \mathcal{E}^{w_\mathcal{E}}(\mathbf{x}_t)) )^2.
\end{aligned}
\end{equation}

Unlike other dimensionality reduction techniques such as AE, PCA, or Diffusion maps~\cite{nadler2006diffusion} that rely on vectorization of input field data, CNN-based approaches consider the spatial structure of the data.
For instance, when an input field is shifted by a pixel, the vectorized version exhibits significant differences, while the convoluted image representation is more robust to such shifts.
By incorporating CNN layers in the autoencoder, our proposed approach benefits from the CNN's ability to capture spatial correlations and exploit the geometric prior encoded in the data.
This allows for a more comprehensive and accurate representation of the underlying structure in the data, which in turn enhances the effectiveness of the dimensionality reduction process.

To capture non-Markovian effects and preserve memory within the low-order manifold (coarse scale), we employ an RNN, a nonlinear propagator, as the fundamental building block of our framework.
The RNN unit learns the forecasting rule as follows:
\begin{gather}
h_t = \mathcal{H}^{w_\mathcal{H}} (\mathbf{z}_t, h_{t- \Delta t}),\\
\mathbf{\tilde{z}}_{t + \Delta t} = \mathcal{R}^{w_\mathcal{R}} (h_t)
\end{gather}
where $h_t \in \mathcal{R}^{d_h}$ is the internal hidden memory state, $\tilde{z}_{t + \Delta t}$ is the latent state prediction, $\mathcal{H}^{w_\mathcal{H}} $ and $\mathcal{R}^{w_\mathcal{R}}$ are the hidden-to-hidden and hidden-to-output mappings with $w_\mathcal{H}$ and $w_\mathcal{R}$ as the respective trainable parameters.
In this study, we use the Long Short-Term Memory (LSTM) implementation of RNN \cite{sherstinsky2020fundamentals,greff2016lstm}, that employs gates to control the information flow and alleviate training problems of previously proposed architectures. 

The role of the RNN in our approach is twofold.
Firstly, it updates its hidden memory state $h_t$ by considering the current input state $\mathbf{z}_t$ and the previous hidden memory state $h_{t- \Delta t}$.
This enables the RNN to effectively track the historical evolution of the low-order state, thereby capturing the non-Markovian dynamics of the system. 
Secondly, leveraging the updated hidden state $h_t$, the RNN forecasts the latent state at the next time-step(s), represented as $\mathbf{\tilde{z}}_{t + \Delta t}$.
Through training, the RNN is optimized to minimize the forecasting loss, quantified by the squared difference between the predicted latent state $\mathbf{\tilde{z}}_{t + \Delta t}$ and the actual latent state at the corresponding future time-step $\mathbf{z}_{t + \Delta t}$.
This optimization process is achieved through BPTT, allowing the RNN to learn and improve its forecasting capabilities over time.

What sets our proposed RefreshNet apart from the LED and other frameworks is its utilization of multiple RNN blocks, hierarchically operating at different timescales.
These timescales increase geometrically by a factor of $k$, thereby capturing varying temporal resolutions.
The initial RNN block, denoted as $\mathcal{R}_1$, operates at a fine-grained time scale of one-step (i.e., $\Delta t_1 = 1$), allowing for precise one-step-ahead predictions.
It is trained to minimize the one-step ahead prediction loss, using training data with a temporal resolution of 1 unit.

Subsequently, higher-level RNN blocks, such as $\mathcal{R}_2$ and $\mathcal{R}_3$, are incorporated into the hierarchy and trained to predict the system dynamics at coarser temporal resolutions.
For instance, $\mathcal{R}_2$ operates with a temporal resolution of $\Delta t_2 = k$, while $\mathcal{R}_3$ operates at an even coarser resolution of $\Delta t_3 = k^2$.
The training data is appropriately subsampled to align with these temporal resolutions.
This hierarchical arrangement of RNN blocks with increasing temporal resolutions enables the RefreshNet to capture multi-scale dynamics and efficiently model system behavior across different time scales.

In order to understand how this proposed hierarchy of RNNs alleviates the effect of error accumulation, it is crucial to consider that the number of prediction steps undertaken directly influences the accumulation of error within the RNN blocks.
Initially, when simulating the system up to the desired point $\mathbf{z}(t)$ with a small number of prediction steps ($s<\epsilon$), the accumulated error remains negligible.
However, as the number of prediction steps increases beyond a threshold ($s> \epsilon$), the accumulated error becomes significant after surpassing $s > \epsilon + C$.
The specific values of $\epsilon$ and $C$ are problem-dependent and rely on various factors, including the system's complexity and the hyperparameters of the RNN, such as the required input sequence length.
The input sequence length signifies the number of past values available to the RNN to predict the subsequent state value.

The hierarchical nature of the RNN building blocks described earlier implies that the system simulation up to the desired point $z(t)$ involves varying levels of error accumulation within each RNN block.
Specifically, when using $\mathcal{R}_1$ with a time step of $\Delta t_1 = 1$, the number of steps required to reach $z(t)$ is $t$.
On the other hand, when using $\mathcal{R}_2$ with a time step of $\Delta t_2 = k$ and $\mathcal{R}_3$ with a time step of $\Delta t_3 = k^2$, the number of steps to reach the same point is reduced to $t/k$ and $t/k^2$, respectively.

Assuming a fixed error per timestep and ignoring the increased difficulty of longer-term predictions, the accumulated error in $\mathcal{R}_3$ is significantly lower compared to that in $\mathcal{R}_2$, which, in turn, is notably lower than the accumulated error in $\mathcal{R}_1$.
This observation stems from the increasing temporal resolutions and the corresponding ability of each RNN block to capture finer details of the system dynamics.
As a result, the hierarchical arrangement of the RNN blocks allows for a more accurate and reliable representation of the system's behavior as we progress through the hierarchy.

Our numerical experiments have revealed an important finding: the value of $\epsilon$ is directly influenced by the selection of the input sequence length.
This observation can be attributed to the intuitive notion that providing a more extended history to the RNN during training and testing enables it to make predictions further into the future.
Consequently, as we increase the input sequence length, the threshold $\epsilon$ also increases, indicating that the accumulated error becomes significant after a greater number of prediction steps.

This finding underscores the critical role of the input sequence length in determining the accuracy and reliability of the predictions made by the RNN.
By considering a more extended history, the RNN can effectively capture the underlying dynamics and improve its forecasting capabilities, thereby reducing the impact of error accumulation.
Therefore, careful consideration and optimization of the input sequence length are crucial for achieving more accurate and longer-term predictions in our framework.
In this context, we set the value of $k$ equal to the input sequence length, which remains the same for all the RNN blocks of the hierarchy.

\begin{figure*}
    \centering
    \includegraphics[width=0.8\textwidth]{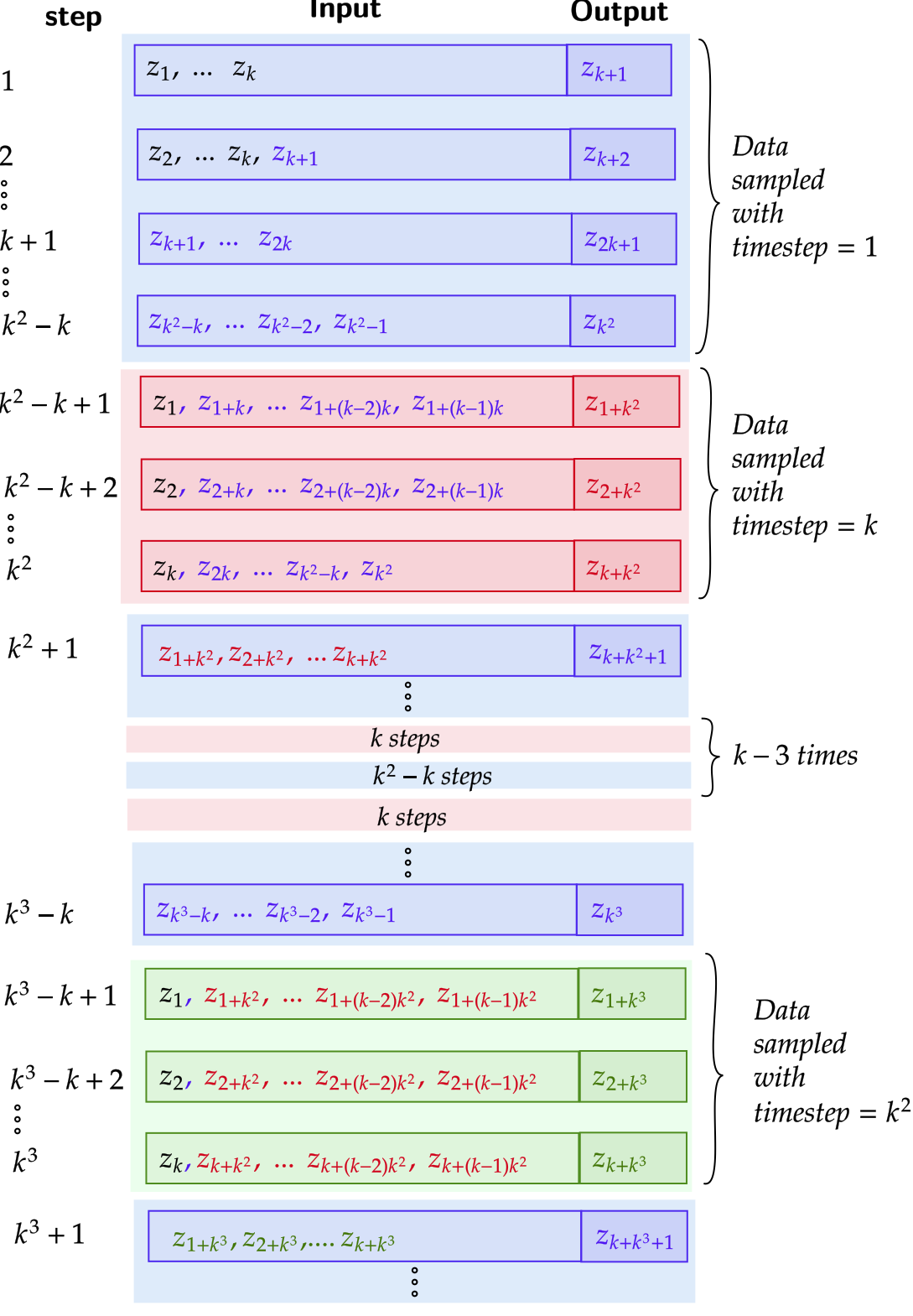}
    \caption{Data sequences at different prediction steps in RefreshNet. Blue, red, and green colors represent the  $\mathcal{R}_1$, $\mathcal{R}_2$, and $\mathcal{R}_3$ models and blue, red, and green fonts represents the data generated by them , respectively. Black fond represents the initial input data sequence. For clearer visualization, see Fig. \ref{data_flow_example} in the Appendix showing a case study for $k=1$.}
    \label{fig:data_flow}
\end{figure*}
 
In addition to capturing multiple scales of dynamics, the hierarchical RNN framework incorporates a refreshing mechanism during autoregressive inference that plays a crucial role in minimizing error accumulation.
Once the first coarser block, $\mathcal{R}_2$, becomes operational, it is employed to refresh the inputs of the finer block, $\mathcal{R}_1$, effectively eliminating the accumulated error within $\mathcal{R}_1$.
The temporal resolution of $\mathcal{R}_2$ is $k$, and its input sequence length is $k$.
As a result, the minimum simulation time required for $\mathcal{R}_2$ to become operational is $t=k^2$.
Similarly, $\mathcal{R}_3$ requires the simulation to have progressed to at least $t=k^3$ to become operational.

\begin{figure*}
    \centering
    \includegraphics[width=\textwidth]{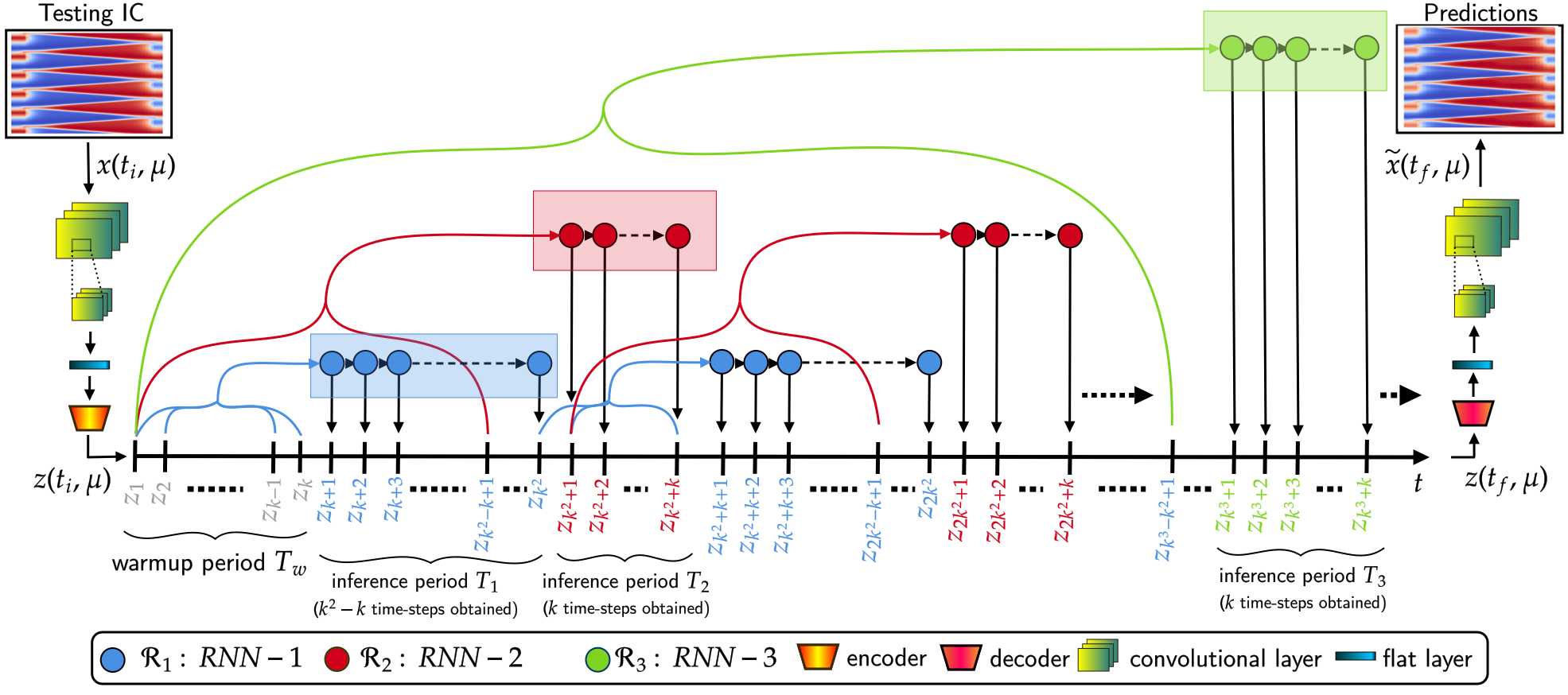}
    \caption{Illustration of RefreshNet architecture. The testing initial condition (IC) is passed through the encoder block to obtain the latent vector $z(t_i, \mu)$ which serves as the input to the different RNN blocks operating with different inference periods. The final predictions are transformed to the original space via the Decoder block.}
    \label{fig:scheme}
\end{figure*}

Once these coarser blocks become operational, they possess freshness and are virtually error-free as they start their first prediction steps, allowing them to refresh the inputs of $\mathcal{R}_1$.
This refreshing process occurs periodically, with $\mathcal{R}_2$ refreshing $\mathcal{R}_1$ every $k^2$ time steps and $\mathcal{R}_3$ refreshing $\mathcal{R}_1$ every $k^3$ time steps.
This cycle continues as we progress through the hierarchy, ensuring that each block performs optimally and minimizes the effective error accumulation.
\Cref{fig:data_flow} shows this process's sequential data flow.
\Cref{fig:scheme} provides a visualization of the entire architecture of RefreshNet.

By incorporating this refreshing mechanism and considering the appropriate simulation length, the hierarchical RNN framework enables accurate and reliable predictions by mitigating the impact of accumulated errors.
Moreover, this mechanism alleviates the exposure bias and the mismatch between training and autoregressive testing and paves the way for forecasting dynamical systems for much longer time scales.

\section{Results}
\label{sec:results}

In this section, we compare RefreshNet with LED and an LSTM across benchmark prototypical applications~\cite{rafiq2023collection}, including the FitzHugh-Nagumo Model, the Reaction-Diffusion equation, and dynamics derived from the Kuramoto-Sivashinsky equation.
All conducted experiments were performed on an Apple M1 Pro, with 16 GB of RAM memory and 8 CPU Cores.
Wherever applicable, the parameters and hyper-parameters were kept the same as LED~\cite{vlachas2022multiscale} to facilitate comparison.

\subsection{FitzHugh-Nagumo Model (FHN)}

RefreshNet is employed to capture the dynamics of the FitzHugh-Nagumo equations (FHN) \cite{fitzhugh1962computation, nagumo1962active}.
The FHN model describes the evolution of an activator \(u(x, t) = \rho_{\text{ac}}(x, t)\) and an inhibitor density \(v(x, t) = \rho_{\text{in}}(x, t)\) on the domain \(x \in [0, L]\):
\begin{align*}
\frac{\partial u}{\partial t} &= D_u \frac{\partial^2 u}{\partial x^2} + u - u^3 - v, \\
\frac{\partial v}{\partial t} &= D_v \frac{\partial^2 v}{\partial x^2} + \varepsilon(u - \alpha_1 v - \alpha_0).
\end{align*}
The system evolves periodically under two timescales, with the activator/inhibitor density acting as the "fast"/"slow" variable, respectively. 
The bifurcation parameter $\varepsilon = 0.006$ controls the difference in the timescales.
The chosen parameter values are: $D_u = 1$, $D_v = 4$, $L = 20$, $\alpha_0 = -0.03$, and $\alpha_1 = 2$, as per \cite{vlachas2022multiscale}.

To discretize the above equations, we utilize a grid with $N = 101$ points and solve them using the Lattice Boltzmann (LB) method \cite{karlin2006elements} with a time-step of $\delta t = 0.005$.
For comparison with the results in \cite{vlachas2022multiscale}, we employ the LB method to generate data from six different initial conditions, obtaining the fine-grained solution used in this study.
The generated data is then sub-sampled, retaining every 200th data point, resulting in a coarse time step of $\Delta t = 1$.
The training set consists of a single time series with 2000 points, the validation set consists of another time series with 2000 points, and the testing set comprises four time series with 10,000 data points each, all originating from different initial conditions. 
The hyperparameters of the AE and LSTM networks are tuned based on the Mean Squared Error (MSE) calculated on the validation data while maintaining consistency with \cite{vlachas2022multiscale} for proper comparison.
The details of the network are provided in Table \ref{fhn_table}.

\begin{table*}
\centering
\caption{Details of the RefreshNet for FHN}
\label{fhn_table}
\begin{tabular}{|c|c|}
\hline
Specifics & Value \\
\hline
Latent Space Generator &  Autoencoder (AE) \\
Number of AE layers &  \{3\} \\
Size of AE layers & \{100\} \\
Activation of AE layers & celu \\
Latent dimension & \{2\} \\
AE Input/Output data scaling & [0,1] \\
AE Output activation & $1+0.5tanh(.)$ \\
AE Weight decay rate & \{0.0\} \\
AE Batch size & 32 \\
AE Initial learning rate & 0.001 \\
RNN cell type & lstm\\
LSTM BPPT sequence length & \{10\} \\
Number of RNN layers in each block & \{1\} \\
Size of RNN layers & \{32\} \\
Activation of RNN Cells & tanh(.) \\
Output activation of RNN Cells & $1+0.5tanh(.)$ \\
\hline
\end{tabular}
\end{table*}

\begin{figure*}
    \centering
    \includegraphics[width=0.7\textwidth]{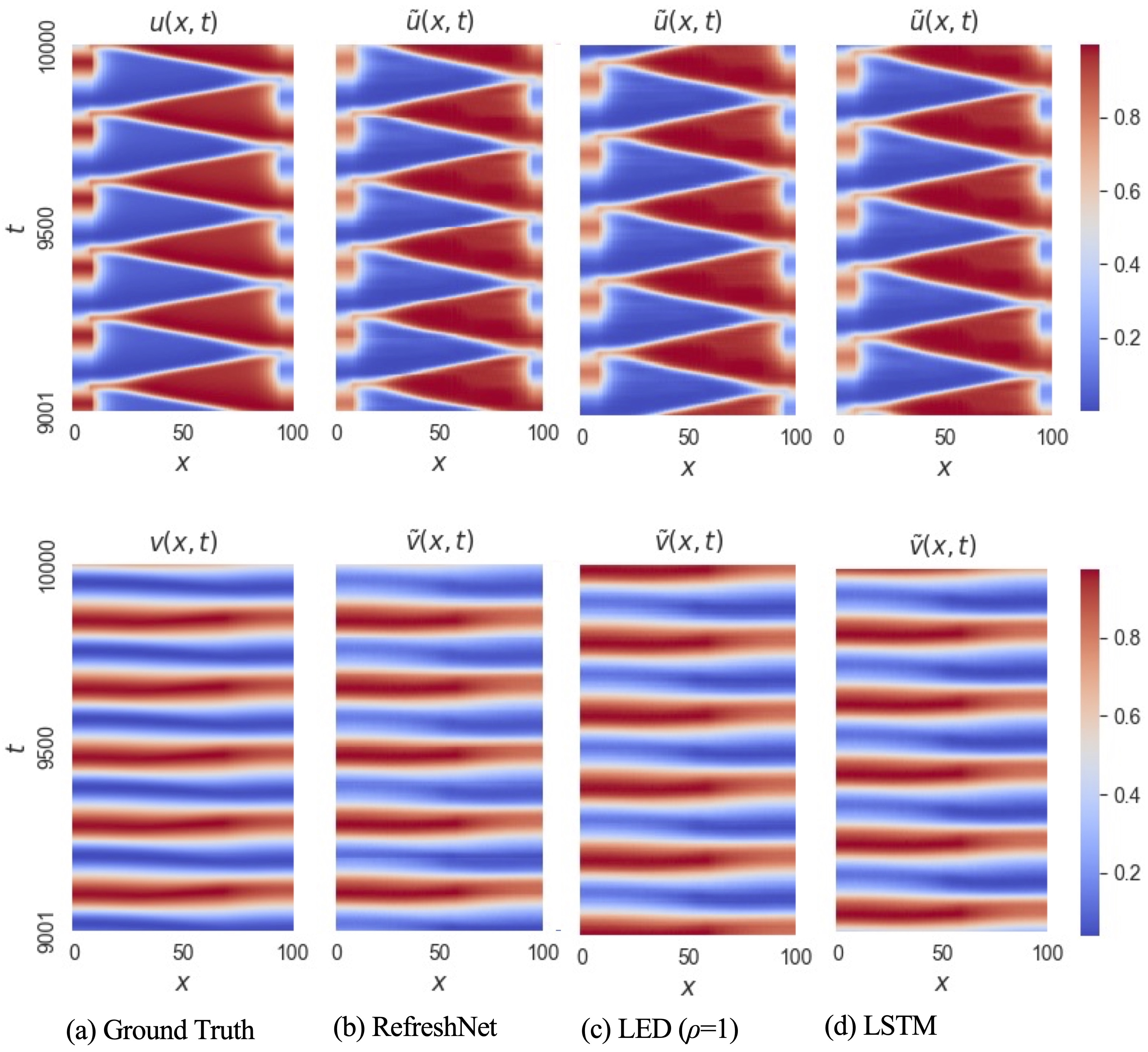}
    \caption{Original trajectories of activator $u(x,t)$ and inhibitor $(v(x,t)$ densities and the corresponding generated trajectories using different methods.}
    \label{fhn_heatmap}
\end{figure*}

\begin{figure}
    \centering
    \includegraphics[width=\columnwidth]{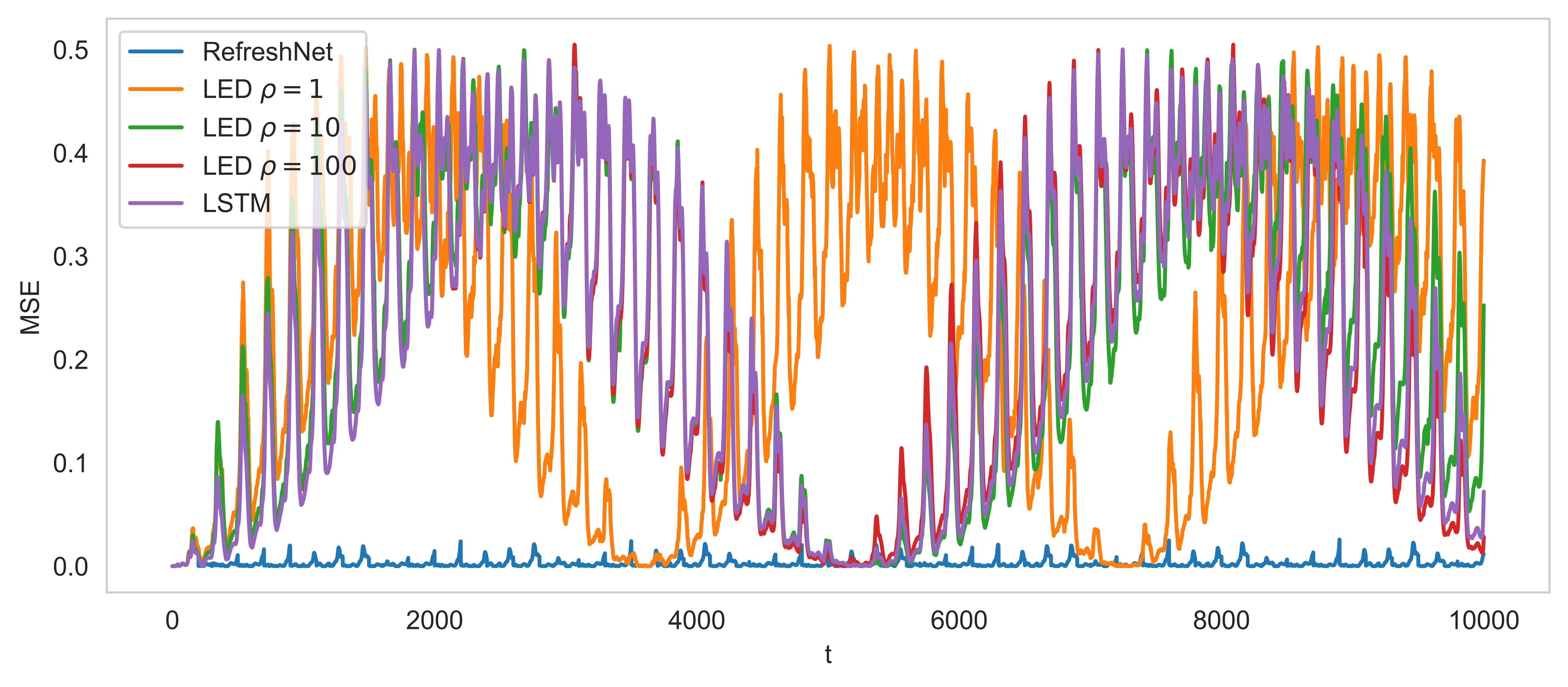}
    \caption{Error accumulation during FHN prediction  in different methods}
    \label{fhnerror}
\end{figure}

\begin{figure}
    \centering
    \includegraphics[width=\columnwidth]{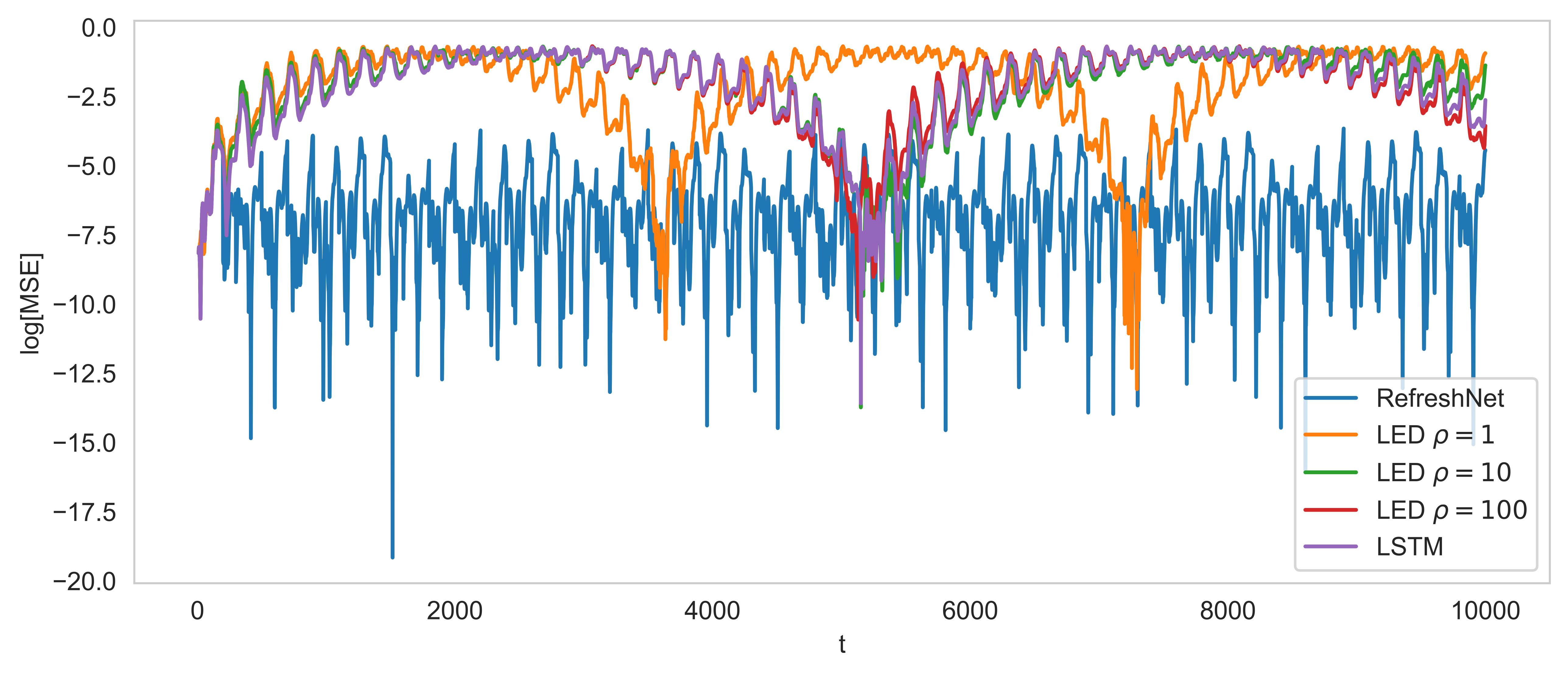}
    \caption{Error accumulation during FHN prediction in different methods (log scale) }
    \label{fhnerrorlog}
\end{figure}

\begin{figure}
    \centering
    \includegraphics[width=\columnwidth]{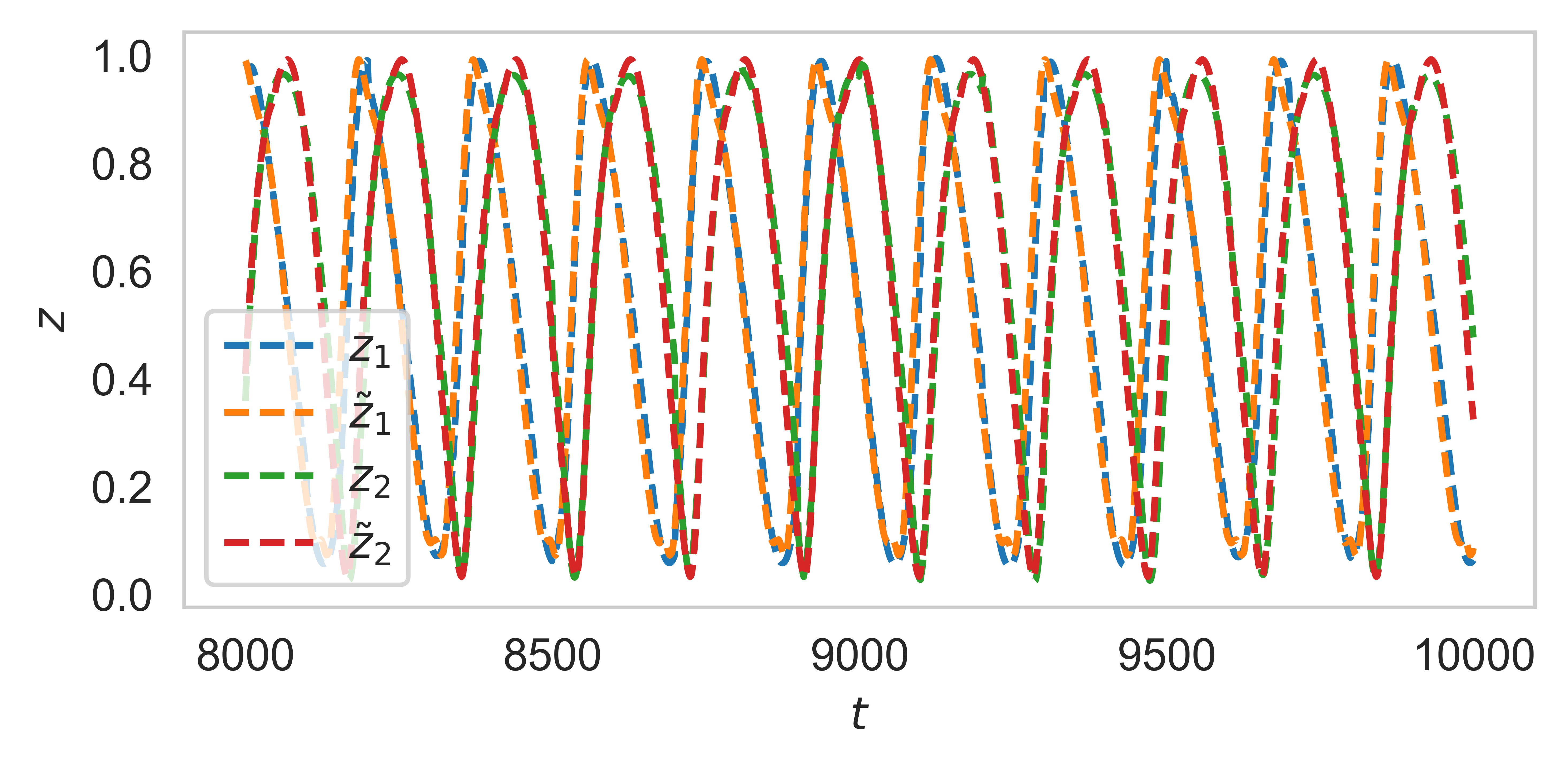}
    \caption{Evolution of FHN latent dynamics }
    \label{FHN_latent_dynamics}
\end{figure}

To compare the performance of our proposed method to LED and LSTM, we consider the MSE.
\Cref{fhn_heatmap} illustrates the results from $t=9,000$ to $t= 10,000$.
The results are comprehensively compared in \Cref{table1}.
The error accumulation is shown in \Cref{fhnerror,fhnerrorlog}.
\Cref{FHN_latent_dynamics} shows the evolution of latent dynamics.
Notably, even at 10,000 prediction steps, the proposed RefreshNet exhibits strikingly minimal error accumulation.
For LED and LSTM, the error accumulates for a certain period and then appears to reduce.
This phenomenon is explained by the periodic nature of the states where the difference in the frequencies and phases of the original and predicted trajectories dominates the error.
The error reduces once the phase difference crosses half of the time-period mark and increases again after a complete cycle.
RefreshNet successfully alleviates the error accumulation problem and demonstrates lower errors than LSTM and the LED.

\begin{table}
\centering
\caption{Accuracy and computational performance compared to the LB method (CPU time: 90.7 s) for FHN model}
\label{table1}
\begin{tabular}{|c|c|c|}
\hline
Method & Computation Time & MSE \\
\hline

LED ($\rho=1$) &  51.10\% & 0.2238 \\
LED ($\rho=10$) &  12.40\% & 0.2296  \\
LED ($\rho=100$) & 3.98\% & 0.2201  \\
LSTM &  2.78\% & 0.2223 \\
RefreshNet &  \textbf{2.43\%} & \textbf{0.0041}  \\
\hline
\end{tabular}
\end{table}

\subsection{Reaction-diffusion equation}

In our investigation, we apply RefreshNet to the lambda-omega reaction-diffusion system \cite{champion2019data,floryan2022data} described by the following equations:
\begin{equation}
\begin{aligned}
\frac{\partial u}{\partial t} &= [1 - (u^2 + v^2)]u + \beta(u^2 + v^2)v + d_1 \nabla^2 u, \\
\frac{\partial v}{\partial t} &= -\beta(u^2 + v^2)u + [1 - (u^2 + v^2)]v + d_2 \nabla^2 v,
\end{aligned}
\end{equation}
where $-10 \leq x, y \leq 10$.
The reaction parameter is $\beta = 1.0$, and the diffusion parameters are $d_1 = d_2 = 1$.
The equations are numerically solved on a $96 \times 96$ uniform grid using the fourth-order Runge-Kutta-Fehlberg method with a $\Delta t = 0.05$ time step.
The system state is represented by the tensor $w = (u, v) \in \mathbb{R}^{2 \times 96 \times 96}$.
Notably, the system exhibits a spiral wave pattern, with the specific shape influenced by the parameter $d$.
The equation is solved using Euler's method with details mentioned in \cite{rafiq2023collection}.

The hyperparameters of the CAE and LSTM networks are tuned based on the Mean Squared Error (MSE) calculated on the validation data while maintaining consistency with \cite{vlachas2022multiscale} for proper comparison.
Hyper-parameters of the network's architecture are provided in Table \ref{rd_table}.

\begin{table*}
\centering
\caption{Details of the RefreshNet for RD}
\label{rd_table}
\begin{tabular}{|c|c|}
\hline
Specifics & Value \\
\hline
Latent Space Generator &  2D convolutional autoencoder \\
Kernels &  Encoder: 5-5-5-5, Decoder: 5-5-5-5 \\
Channels & 2-16-16-16-16-8-16-16-16-16-2 \\
Activation of CNN layers & celu \\
Latent dimension & \{8\} \\
CAE Input/Output data scaling & [0,1] \\
CAE Output activation & $1+0.5tanh(.)$ \\
CAE Weight decay rate & \{0.0\} \\
CAE Batch size & 32 \\
CAE Initial learning rate & 0.001 \\
RNN cell type & lstm\\
LSTM BPPT sequence length & \{10\} \\
Number of RNN layers in each block & \{1\} \\
Size of RNN layers & \{64\} \\
Activation of RNN Cells & tanh(.) \\
Output activation of RNN Cells & $1+0.5tanh(.)$ \\
\hline
\end{tabular}
\end{table*}

\begin{figure*}
    \centering
    \includegraphics[width=\textwidth]{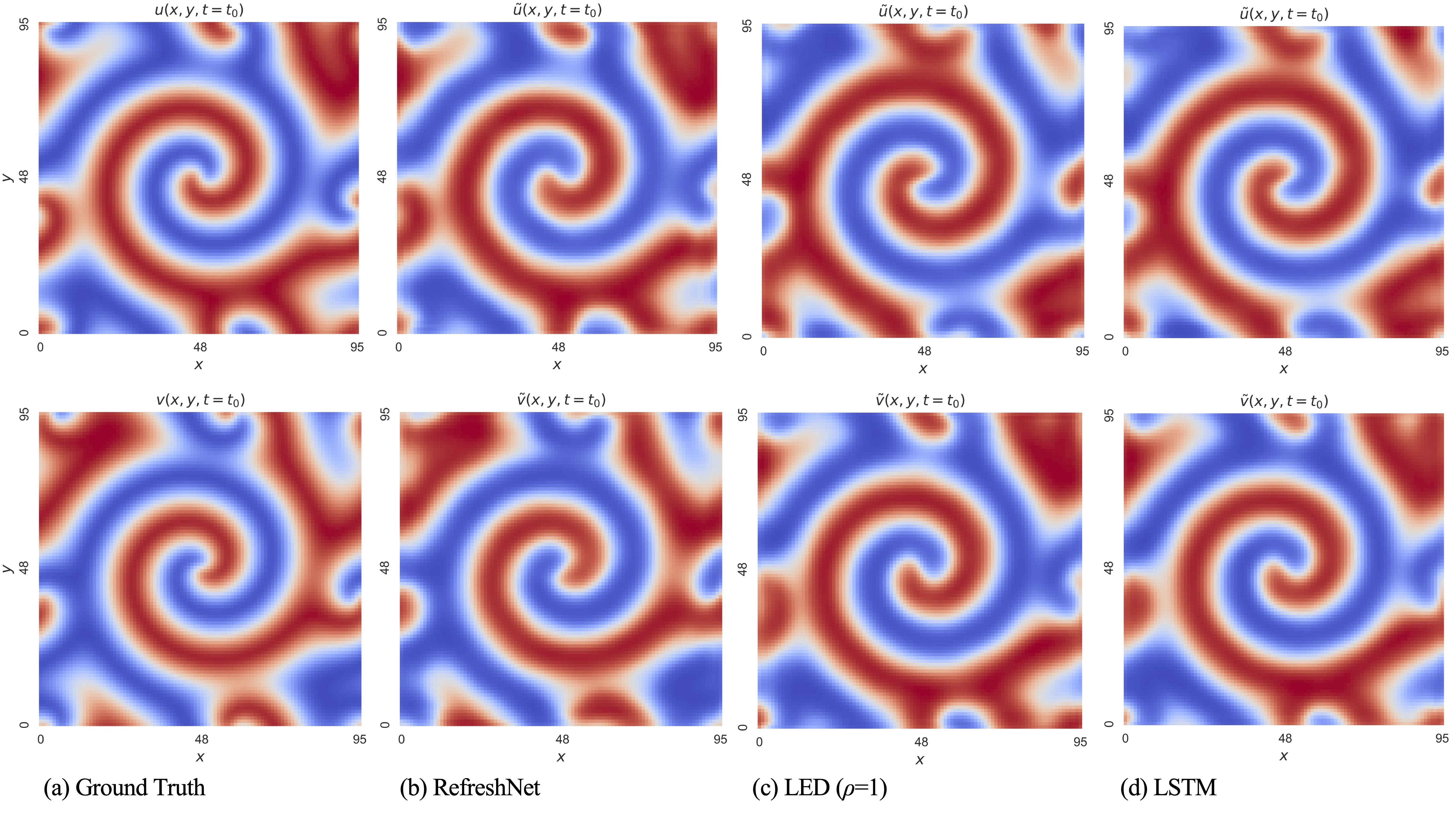}
    \caption{Original trajectories of RD equation and the corresponding generated trajectories using different methods at $t=10,000$.}
    \label{RD_heatmaps}
\end{figure*}

\begin{figure*}
    \centering
    \includegraphics[width=\textwidth]{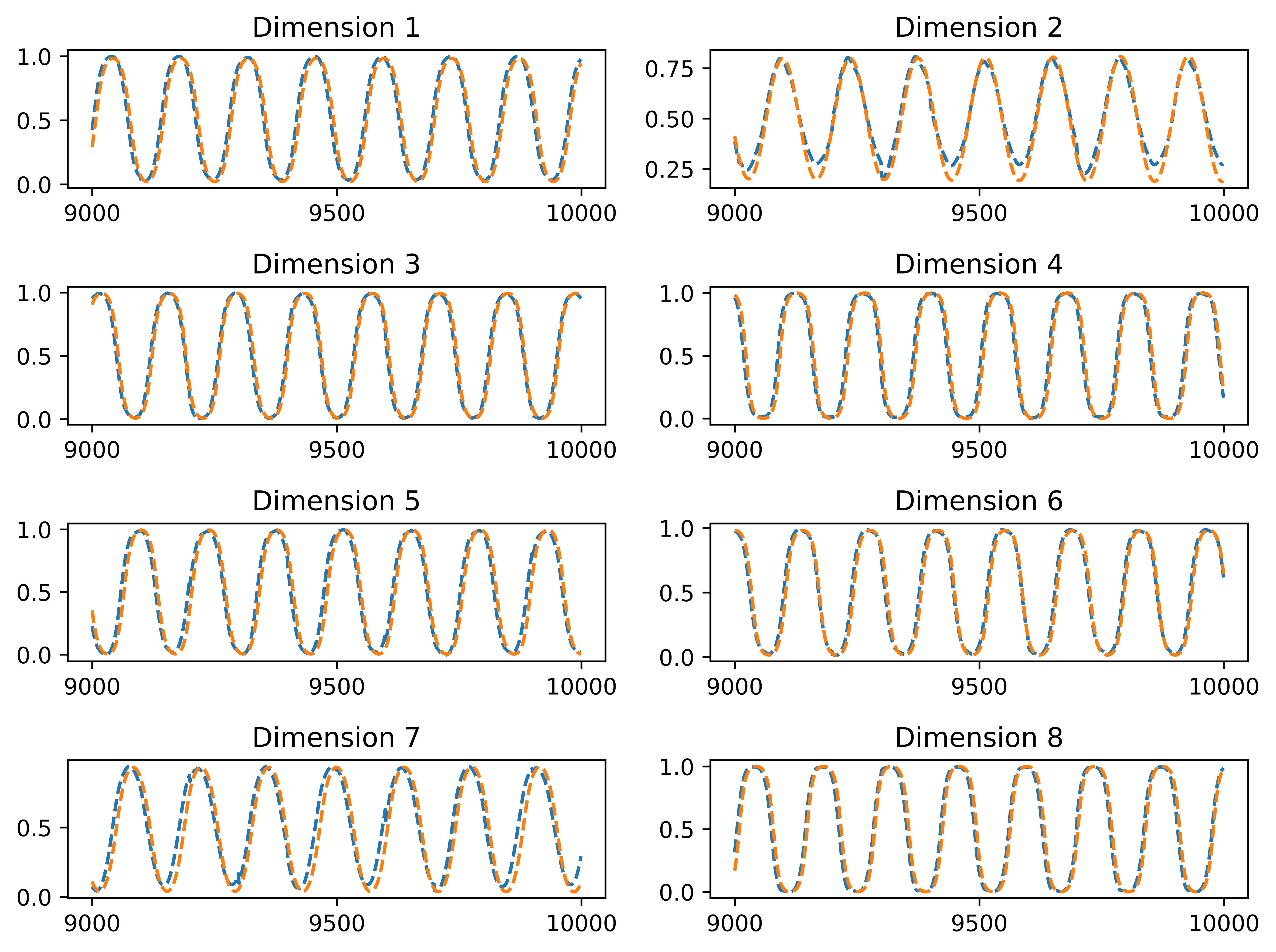}
    \caption{Evolution of latent dynamics in RD model using RefreshNet. Each subplot shows the original and the predicted values.}
    \label{RD_latent_dynamics}
\end{figure*}

\begin{figure}
    \centering
    \includegraphics[width=\columnwidth]{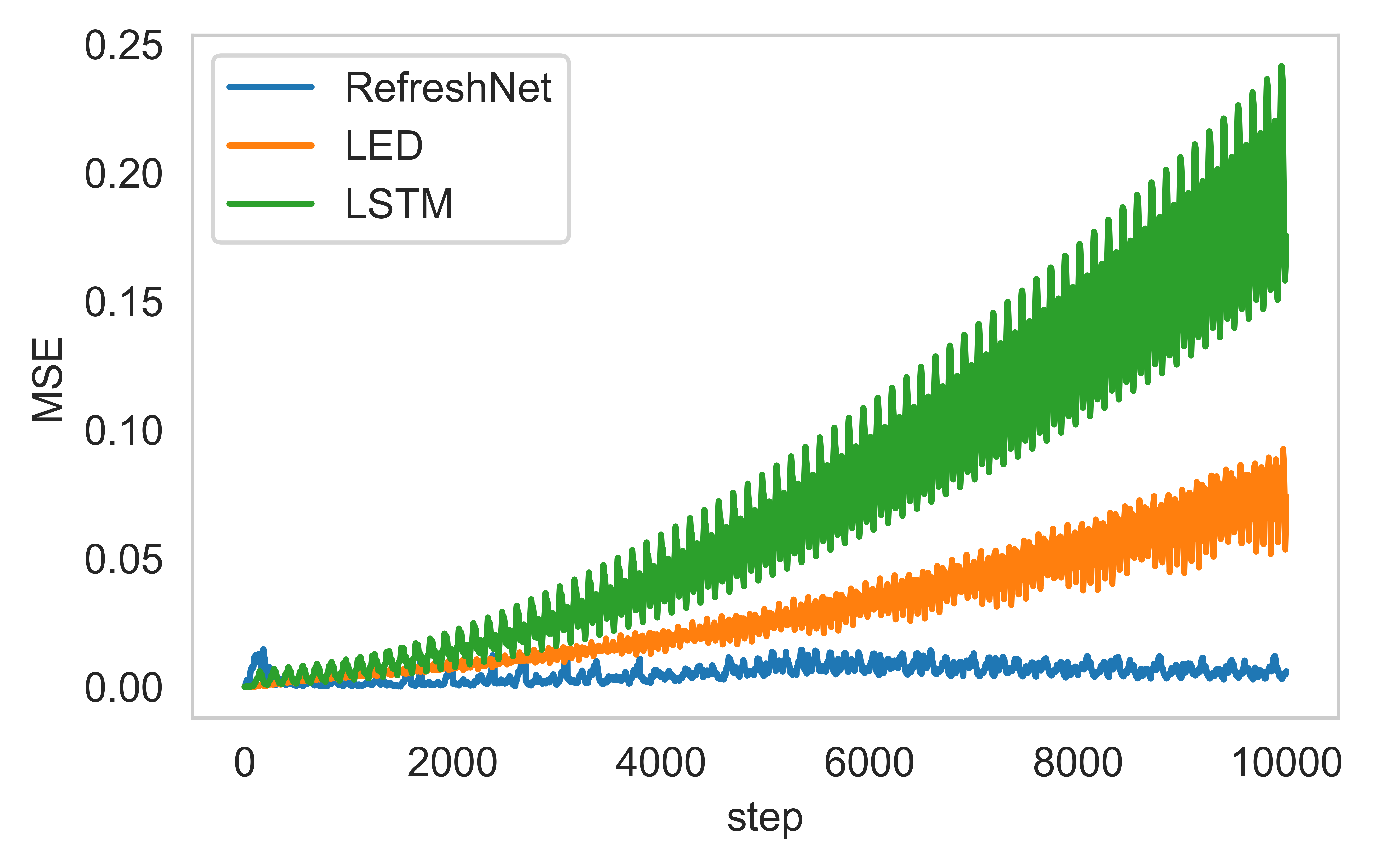}
    \caption{Error accumulation during RD prediction  in different methods}
    \label{RD_error}
\end{figure}

\begin{figure}
    \centering
    \includegraphics[width=\columnwidth]{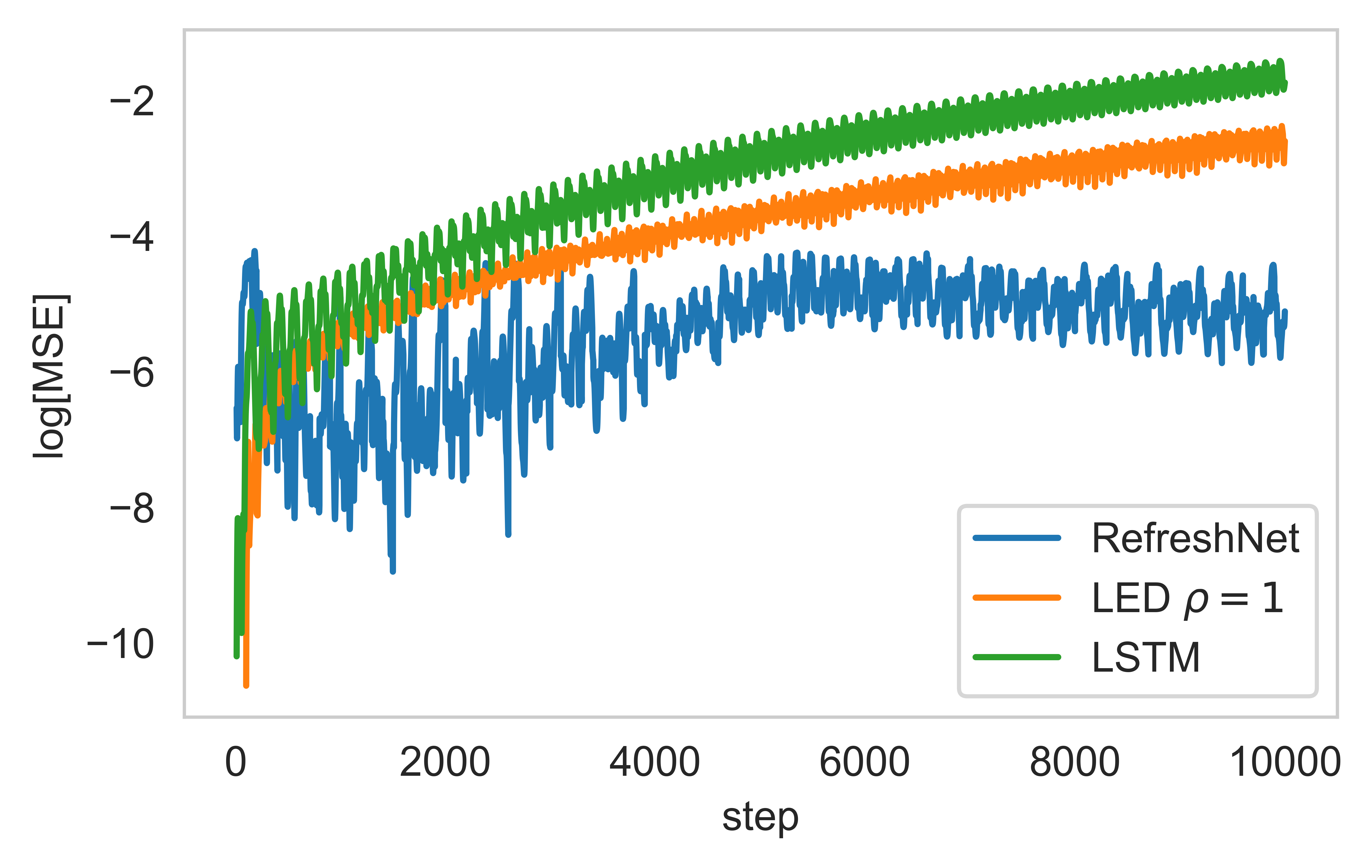}
    \caption{Error accumulation during RD prediction in different methods (log scale) }
    \label{RD_error_log}
\end{figure}

To assess the performance of RefreshNet, we employ the mean squared error (MSE) as the metric to measure the error. 
The simulation is conducted for 10,000 time steps. 
\Cref{RD_heatmaps} illustrates the results of our simulations at $t = 10,000$ in comparison to the LSTM and the LED.
Table \ref{table2} presents a comprehensive comparison of the results.
The error accumulation is shown in \Cref{RD_error,RD_error_log}. 
\Cref{RD_latent_dynamics} shows the evolution of latent dynamics.
Remarkably, the proposed RefreshNet exhibits minimal error accumulation, confirming its effectiveness.

\begin{table}
\centering
\caption{Accuracy and computational performance of different methods for RD model compared to the original solution \cite{rafiq2023collection} (CPU time: 203.0 s))}
\label{table2}
\begin{tabular}{|c|c|c|}
\hline
Method & Computation Time & MSE \\
\hline

LED ($\rho=1$) &  52.1\% & 0.07 \\
LSTM &  1.8\% & 0.2692 \\
RefreshNet &  \textbf{1.9\%} & \textbf{0.0054}  \\
\hline
\end{tabular}
\end{table}

\subsection{Kuramoto-Sivashinsky}

The Kuramoto-Sivashinsky (KS) \cite{kuramoto1978diffusion, ashinsky1988nonlinear} is a prototypical partial differential equation (PDE) of fourth order that exhibits a very rich range of nonlinear phenomena, and is frequently used as a benachmark of chaotic dynamics~\cite{vlachas2018data,vlachas2020backpropagation, pathak2018model,pathak2017using}.
In case of high dissipation and small spatial extent $L$ (domain size), the long-term dynamics of KS can be represented on a low-dimensional inertial manifold \cite{robinson1994inertial,linot2020deep}, that attracts all neighboring states at an exponential rate after a transient period.
Here, RefreshNet and LED are employed to learn the low-order manifold of the effective dynamics in KS and forecast their long-term evolution.

The one-dimensional K-S equation is given by the PDE:
\begin{equation}
\frac{\partial u}{\partial t} = -\nu \frac{\partial^4 u}{\partial x^4} - \frac{\partial^2 u}{\partial x^2} - u \frac{\partial u}{\partial x}.
\label{eq:ks}
\end{equation}
on the domain $\Omega = [0, L]$ with periodic boundary conditions $u(0, t) = u(L, t)$ and $\nu = 1$.
The special case $L = 22$ considered in this work is studied extensively in \cite{cvitanovic2010state} and exhibits a structurally stable chaotic attractor, i.e., an inertial manifold where the long-term dynamics lie.
\Cref{eq:ks} is discretized with a grid of size 64 points and solved using the fourth-order method for stiff PDEs introduced in \cite{kassam2005fourth} with a time-step of $\delta t = 2.5 \times 10^{-3}$ starting from a random initial condition.
The data is subsampled to $\Delta t = 0.25$ (coarse time-step of RefreshNet and LED).
$15 \times 10^3$ samples are used for training, and another $15 \times 10^3$ for validation.
For testing purposes, the process is repeated with a different random seed, generating another $15 \times 10^3$ samples.

The hyperparameters of the CAE and LSTM networks (components of the RefreshNet and LED frameworks) are tuned based on the Mean Squared Error (MSE) calculated on the validation data while maintaining consistency with \cite{vlachas2022multiscale} for proper comparison.
The details of the network are provided in Table \ref{ks_table}.

\begin{table*}
\centering
\caption{Details of the RefreshNet for KS}
\label{ks_table}
\begin{tabular}{|c|c|}
\hline
Specifics & Value \\
\hline
Latent Space Generator &  1D convolutional autoencoder \\
Kernels &  Encoder: 5-5-5-5, Decoder: 5-5-5-5 \\
Channels & 1-16-32-64-8-8-8-64-32-16-1 \\
Activation of CNN layers & celu \\
Latent dimension & \{8\} \\
CAE Input/Output data scaling & [0,1] \\
CAE Output activation & $1+0.5tanh(.)$ \\
CAE Weight decay rate & \{0.0\} \\
CAE Batch size & 32 \\
CAE Initial learning rate & 0.001 \\
RNN cell type & lstm\\
LSTM BPPT sequence length & \{10\} \\
Number of RNN layers in each block & \{1\} \\
Size of RNN layers & \{512\} \\
Activation of RNN Cells & tanh(.) \\
Output activation of RNN Cells & $1+0.5tanh(.)$ \\
\hline
\end{tabular}
\end{table*}

\begin{figure}
    \centering
    \includegraphics[width=\columnwidth]{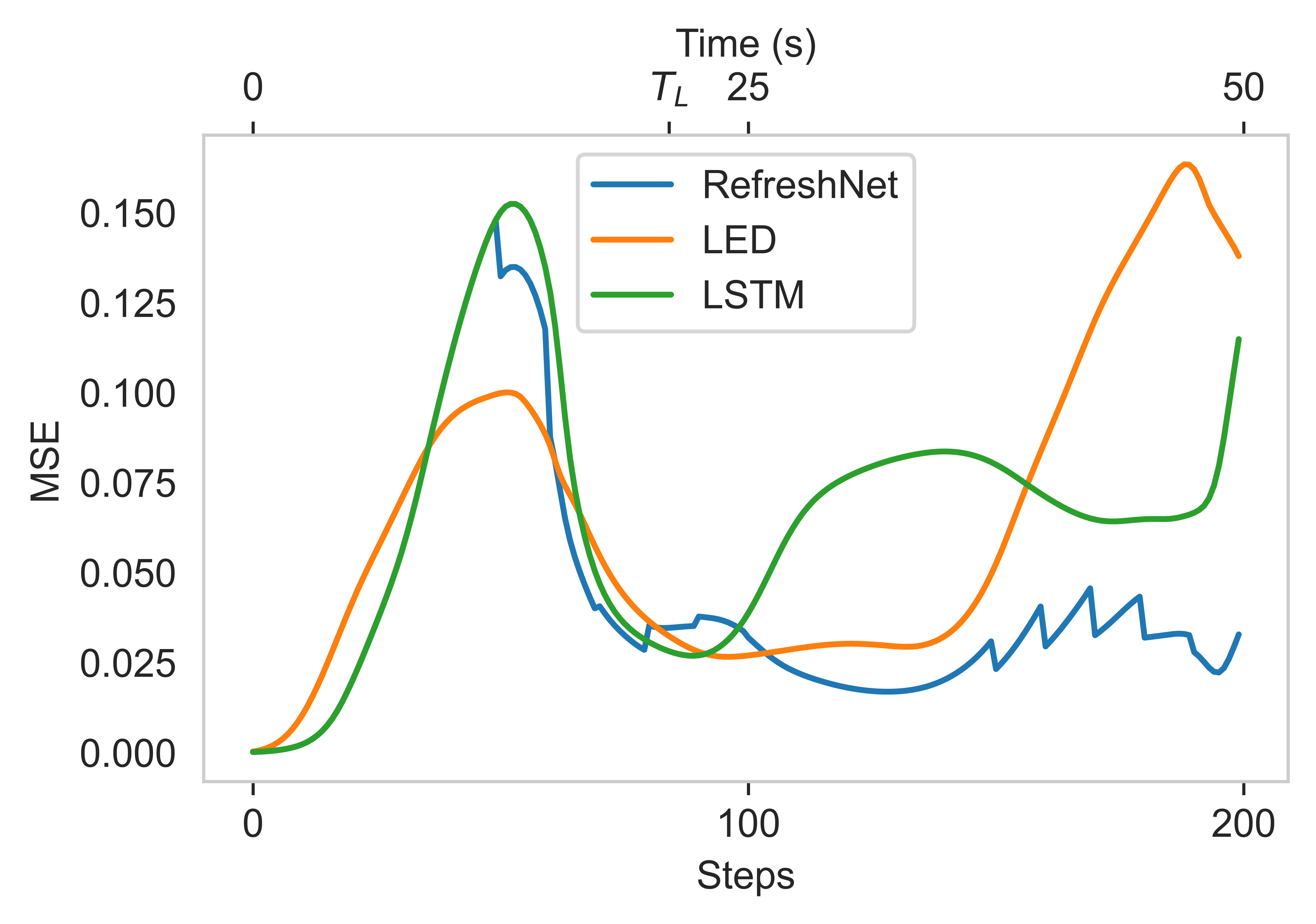}
    \caption{Error accumulation during KS prediction using different methods ($T_L = 20.83$ is the Lyapunov time)}
    \label{KS_error}
\end{figure}

\begin{figure}
    \centering
    \includegraphics[width=\columnwidth]{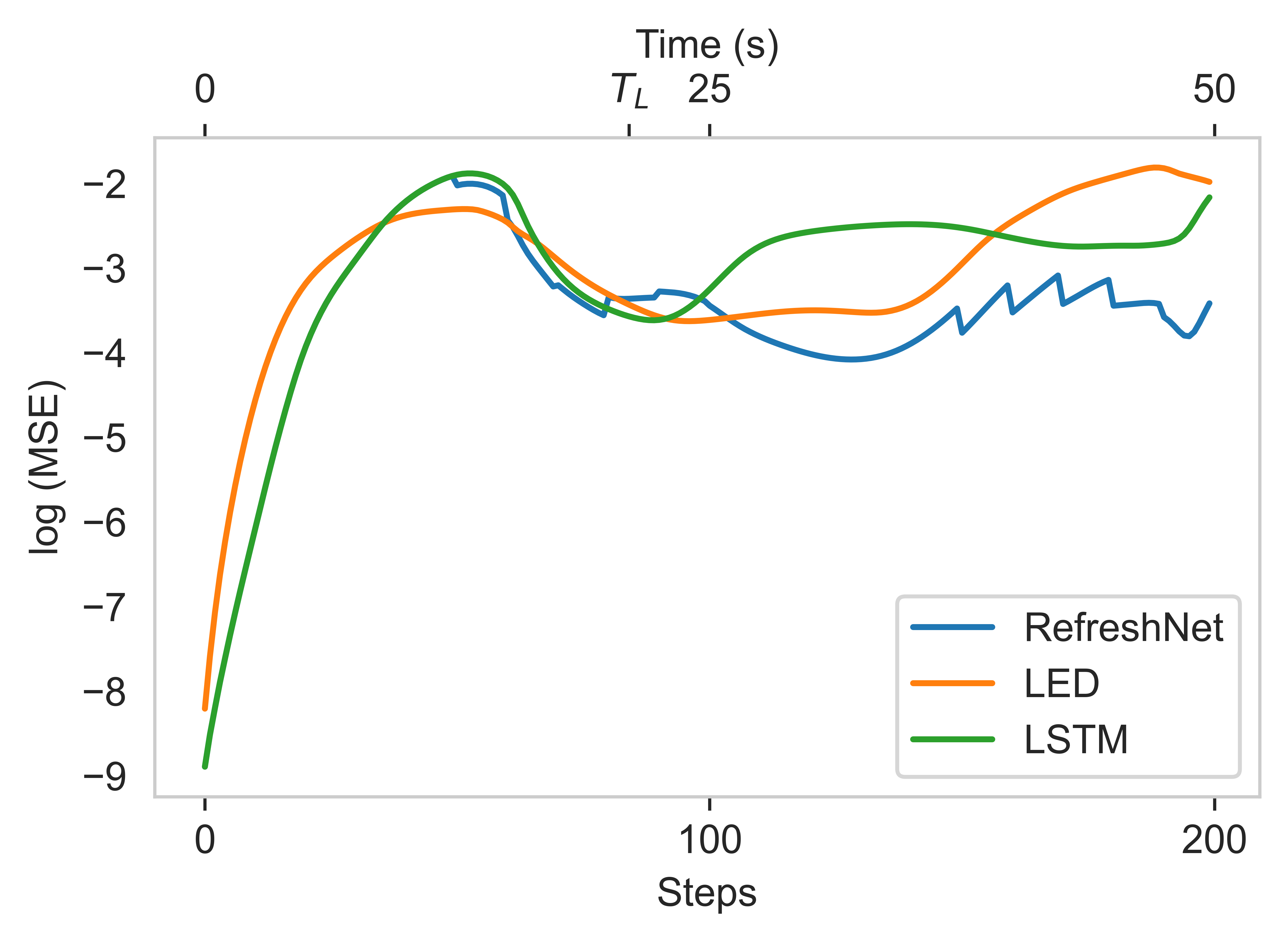}
    \caption{Error accumulation (log scale) during KS prediction using different methods ($T_L = 20.83$ is the Lyapunov time) }
    \label{KS_error_log}
\end{figure}

\begin{figure*}
    \centering
    \includegraphics[width=0.7\textwidth]{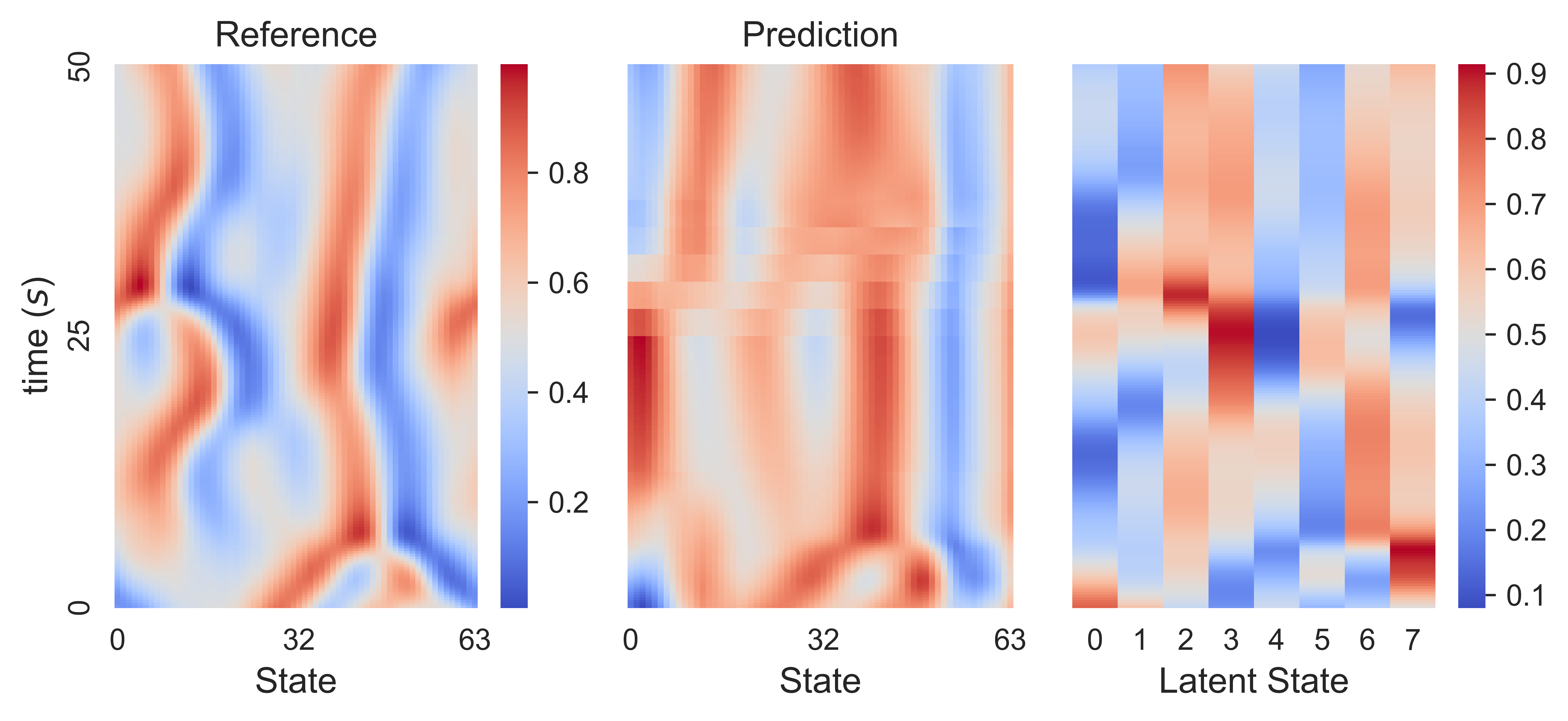}
    \caption{Evolution of reference, predicted, latent dynamics in KS model using RefreshNet}
    \label{KS_heatmap}
\end{figure*}

To compare the performance of our proposed method with LED, we consider the MSE. \Cref{KS_error,KS_error_log,KS_heatmap} qualitatively illustrate the results up to $t=50$, consistent with \cite{vlachas2022multiscale}.
Notably, the maximum Lyapunov exponent of the KS equation for spatial extend $L=22$ is approximately $0.043$~\cite{edson2019lyapunov}, implying a Lyapunov time of $23.25$ unit times.
We observe that prediction after this time deteriorates, which is expected due to the chaoticity of the system, as even minimal prediction errors close to machine precision accumulate exponentially.
Table \ref{table3} provides a comprehensive comparison of the results. 
RefreshNet demonstrates a $41\%$ reduction in Mean Squared Error (MSE) compared to LSTM, with a similar computational cost, highlighting its significant efficiency gains.
Compared to LED, RefreshNet achieves a $30\%$ improvement in MSE and operates an order of magnitude faster.


\begin{table}
\centering
\caption{Accuracy and computational performance of different methods for KS model compared to the numerical simulation \cite{kassam2005fourth} (CPU time: 61.0 s)}
\label{table3}
\begin{tabular}{|c|c|c|}
\hline
Method & Computation Time & MSE \\
\hline

LED ($\rho=0.25$) &  83.1\% & 0.07410 \\
LSTM &  2.1\% & 0.08916 \\
RefreshNet &  \textbf{2.3\%} & \textbf{0.0522}  \\
\hline
\end{tabular}
\end{table}

\section{Discussion}
\label{sec:discussion}

The hierarchical multi-timescale RNN framework proposed in this study presents a unique and powerful approach to capturing the dynamics of complex systems with multiple time scales.
By incorporating multiple RNN blocks at different temporal resolutions, the framework captures fine-grained details and takes advantage of coarse-grained information to refresh and refine predictions.

This refreshing mechanism is a crucial feature that sets our framework apart from traditional RNN models.
The coarser blocks' ability to refresh the finer blocks' inputs effectively resets the error accumulation, leading to more accurate and reliable predictions.
This process allows the framework to achieve a delicate balance between capturing intricate details and minimizing the impact of accumulated errors, resulting in superior performance compared to conventional RNN models.

Furthermore, the hierarchical nature of the framework provides a natural hierarchy of scales, with each RNN block operating at a different temporal resolution.
This hierarchical structure enables the framework to adaptively capture the dynamics at various scales, ranging from fine-scale fluctuations to broader trends.
By incorporating multiple scales in the modeling process, our framework offers a more comprehensive understanding of complex systems and their evolution.

In contrast to the state-of-the-art LED method, the proposed framework alleviates error accumulation and enables much larger prediction times.
Due to the error accumulation, the LED framework would resort to simulating the original high-dimensional dynamics early on, increasing its computational cost and time to solution.
In contrast, the proposed framework leverages the capability of a hierarchy of multi-timescale RNN blocks of learning the dynamics and refreshing the prediction in autoregressive inference, alleviating the need for explicit numerical simulations and thus achieving significant reductions in computational time.
This reduction in computation time is highly advantageous, especially when dealing with large-scale complex systems or performing real-time predictions.

The benefits of the hierarchical RNN framework extend beyond accurate prediction.
The framework's modular and scalable nature allows for easy integration of additional RNN blocks at even coarser temporal resolutions, enabling dynamics modeling across a broader range of scales.
This flexibility and scalability make our framework well-suited for studying complex systems with evolving dynamics, such as climate systems, financial markets, and biological processes.

By combining accuracy on long-term forecasting and scalability, our proposed framework surpasses LED and other existing methods in capturing the long-term dynamics of complex multiscale systems.
It offers an innovative and efficient approach that can unlock new possibilities for studying and understanding complex phenomena across various domains. 

By leveraging convolutional autoencoders and recurrent neural networks, our framework captures the intricate dynamics of complex systems and learns a compressed latent space representation by LED.
This latent space, encoding the essential features and patterns of the complex system, holds the potential for generative modeling applications, expanding our framework's potential beyond traditional forecasting tasks.
It enables us to explore creative possibilities, such as scenario generation, data augmentation, and synthetic data generation for training purposes. 
The ability to generate new samples that capture the essence of the complex system's dynamics contributes to a more comprehensive understanding of the system and facilitates hypothesis testing and scenario analysis. 
RefreshNet establishes a unique fusion of data-driven methods and first-principles models, unlocking the potential for precise and resource-efficient prediction of complex multiscale systems. 
This framework can be applied to problems in which data is generated from first principles or collected from sensors.

While our proposed framework offers significant advancements and numerous benefits, there are certain limitations to consider and areas for future research.
Firstly, the reliance on the latent space obtained through convolutional autoencoders assumes that it adequately represents the dynamics of the complex system.
Exploring alternative generative modeling approaches, such as variational autoencoders or generative adversarial networks, can enhance representation capabilities.
Additionally, extending the framework to consider external influences or perturbations would enhance its applicability to a broader range of systems.
Incorporating external inputs and developing methods to integrate exogenous factors can improve the framework's robustness and adaptability.
Handling data limitations, such as missing or noisy data, can be addressed through data imputation or noise reduction approaches while incorporating domain knowledge and expert guidance can enhance the framework's performance.
RefreshNet is designed as a method that operates across multiple timescales.
Its capabilities can be enhanced through integration with multiscale autoencoders~\cite{liu2023multiresolution}, enabling it to effectively tackle problems that have multiscale characteristics in the spatial domain.

\section{Conclusion}
\label{sec:conclusion}

In this paper, we have presented a data-driven deep learning model called RefreshNet that effectively captures the dynamics of complex systems across multiple temporal resolutions.
By incorporating convolutional autoencoders and multiple recurrent neural network (RNN) blocks operating at geometrically increasing timescales, our framework leverages the power of fine-grained details and coarse-grained information to improve long-term predictions of complex system dynamics.

We have demonstrated that the refreshing mechanism in our framework, where coarser blocks refresh the inputs of finer blocks, plays a crucial role in resetting error accumulation and improving long-term prediction accuracy.
This feature sets our framework apart from traditional RNN models and the LED framework and contributes to its superior performance in capturing complex system dynamics.

Through experiments and comparisons with state-of-the-art techniques such as LED on three applications, namely the FitzHugh Nagumo system, the Reaction-Diffusion equation, and the Kuramoto-Sivashinsky dynamics, we have shown that our framework effectively alleviateds the issue of error accumulation, leading to more reliable and accurate predictions.
These results demonstrate the potential of the RefreshNet framework to advance the field of complex system modeling.

The hierarchical nature of our framework enables the adaptive modeling of dynamics at different scales, capturing both fine-scale fluctuations and broader trends.
This flexibility is further enhanced by our framework's modularity and scalability, allowing for the easy integration of additional RNN blocks at even coarser temporal resolutions. 
As a result, our framework is well-suited for studying complex systems with evolving dynamics. 
Moreover, in contrast to LED, alleviating the need to numerically solve the original equations, which entail an exhaustive computational burden, drastically reduces the computation time for our framework.

The proposed RefreshNet framework opens up exciting possibilities for further research and applications in various domains.
Its ability to capture complex system behavior across multiple scales offers a more comprehensive understanding of these systems and their evolution.
This framework has the potential to contribute to advancements in fields such as climate science, financial markets, and biological processes, where the accurate modeling of complex dynamics is crucial.

In conclusion, the hierarchical refresh framework for multiscale learning presented in this paper represents a significant advancement in capturing the dynamics of complex systems.
By leveraging convolutional autoencoders, multiple RNN blocks, and the refreshing mechanism, our framework offers improved prediction accuracy, eliminates error accumulation, and enables the modeling of dynamics at various resolutions.
We believe this framework will inspire further research in the field and find valuable applications in understanding and predicting the behavior of complex systems. 
The results for the KS model suggest RefreshNet's utility in modeling chaotic systems, a potential avenue we plan to explore in the future.
\section*{Appendix}
\begin{figure*}
    \centering
    \includegraphics[width=0.8\textwidth]{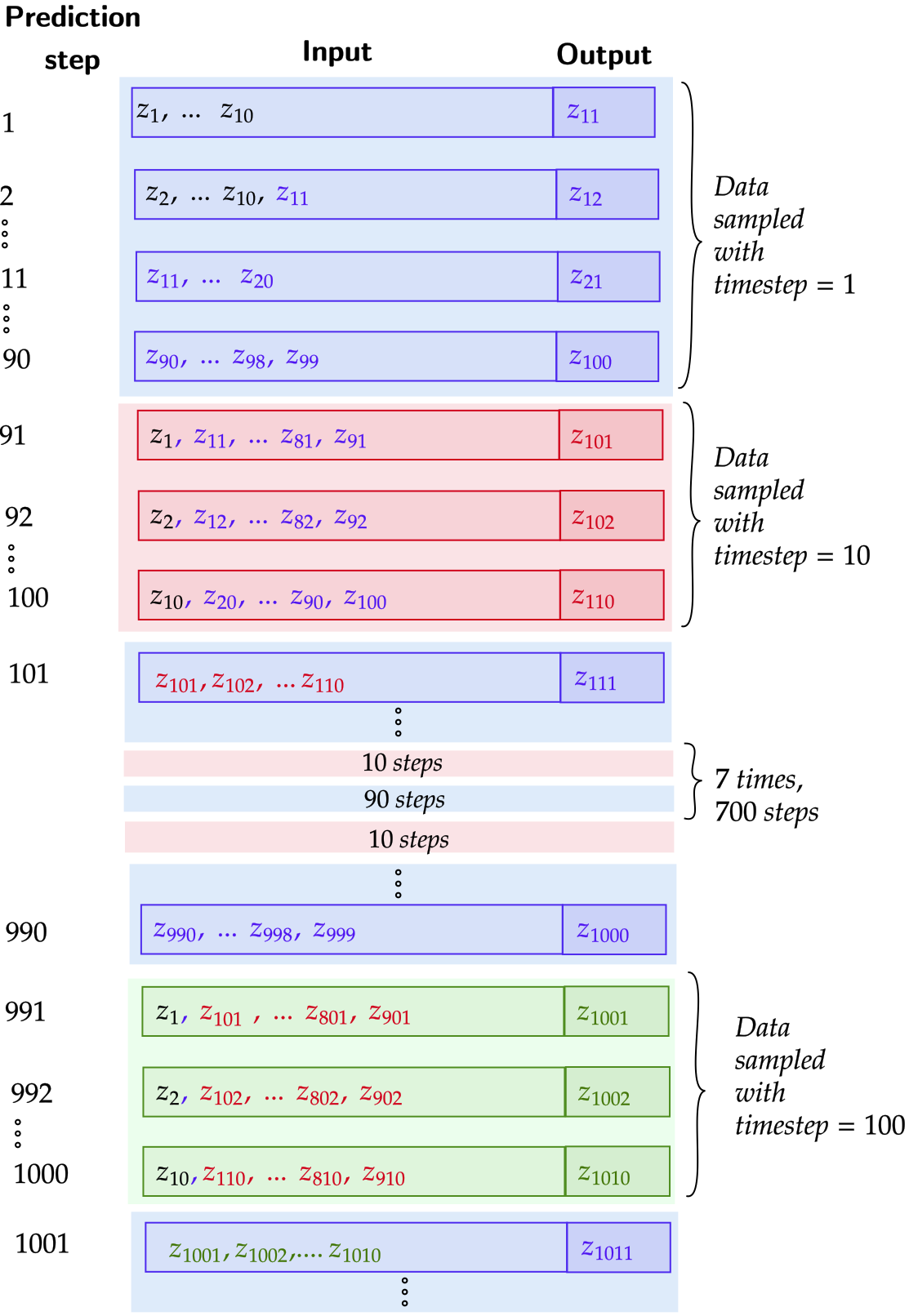}
    \caption{Data sequences at different prediction steps in RefreshNet for the case $k=1$. Blue, red, and green colors represent the  $\mathcal{R}_1$, $\mathcal{R}_2$, and $\mathcal{R}_3$ models and blue, red, and green fonts represents the data generated by them , respectively. Black fond represents the initial input data sequence.}
    \label{data_flow_example}
\end{figure*}

Here, a data sequences for RefreshNet are shown for a case of $k=10$ in Fig. \ref{data_flow_example}

\backmatter

%
%
%

\section*{Data availability}
The datasets generated during and/or analysed during the current study are not publicly available  but are available from the corresponding author on reasonable request.

\section*{Declarations}

\begin{itemize}
\item Funding: First author acknowledges the funding from the Ministry of Education, Government of India. Second author acknowledges the funding from the Science and Engineering Research Board (SERB), Department of Science and Technology (DST), Government of India via Grant No. PDF/2022/002081. 
\item Conflict of interest: The authors declare no conflicts of interests.
\end{itemize}

%
%
%
%
%
%
%
%
%


\bibliography{bibliography}

\end{document}